\documentclass[runningheads]{llncs}
\usepackage[T1]{fontenc}
\usepackage{cite}
\usepackage{graphicx}
\usepackage{amsmath}
\usepackage{booktabs}
\usepackage{array}
\usepackage[colorinlistoftodos,prependcaption,textsize=tiny]{todonotes}
\begin{document}
%
\title{PHOCUS: Physics-Based Deconvolution for Ultrasound Resolution Enhancement}
%
\author{Felix Duelmer \inst{1,2,3,4}\thanks{Corresponding Author} \and
Walter Simson \inst{5} \and
Mohammad Farid Azampour \inst{1,2} \and Magdalena Wysocki\inst{1, 2, 6} \and Angelos Karlas \inst{7, 8} \and Nassir Navab\inst{1, 2}}
\authorrunning{F. Duelmer et al.}
%

%
\institute{
Chair for Computer-Aided Medical Procedures and Augmented Reality, School of Computation, Information and Technology, Technical University of Munich, Munich, Germany \and
Munich Center for Machine Learning (MCML), Munich, Germany \and
Institute of Biological and Medical Imaging, Helmholtz Zentrum Munich, Neuherberg, Germany \and
Chair of Biological Imaging, School of Medicine, Technical University of Munich, Munich, Germany \and
Department of Radiology, School of Medicine, Stanford University, Stanford,  USA \and
LUMA Vision GmbH, Munich, Germany \and
Department for Vascular and Endovascular Surgery, Rechts der Isar University Hospital, Technical University of Munich, Munich, Germany \and
German Centre for Cardiovascular Research, Munich, Germany}

\maketitle              
\begin{abstract}
Ultrasound is widely used in medical diagnostics allowing for accessible and powerful imaging but suffers from resolution limitations due to diffraction and the finite aperture of the imaging system, which restricts diagnostic use. The impulse function of an ultrasound imaging system is called the point spread function (PSF), which is convolved with the spatial distribution of reflectors in the image formation process. Recovering high-resolution reflector distributions by removing image distortions induced by the convolution process improves image clarity and detail. Conventionally, deconvolution techniques attempt to rectify the imaging system's dependent PSF, working directly on the radio-frequency (RF) data. However, RF data is often not readily accessible. Therefore, we introduce a physics-based deconvolution process using a modeled PSF, working directly on the more commonly available B-mode images. By leveraging Implicit Neural Representations (INRs), we learn a continuous mapping from spatial locations to their respective echogenicity values, effectively compensating for the discretized image space. 
Our contribution consists of a novel methodology for retrieving a continuous echogenicity map directly from a B-mode image through a differentiable physics-based rendering pipeline for ultrasound resolution enhancement. We qualitatively and quantitatively evaluate our approach on synthetic data, demonstrating improvements over traditional methods in metrics such as PSNR and SSIM. Furthermore, we show qualitative enhancements on an ultrasound phantom and an in-vivo acquisition of a carotid artery. 

\end{abstract}
\section{Introduction}

Medical ultrasound (US) imaging examines biological tissue using acoustic sound waves to create images based on the received echoes.
The relatively low frequency and limited aperture size compared to other imaging modalities like CT and MRI reduce the diffraction-limited resolution of ultrasound imaging. 
This limitation affects the diagnostic efficacy of ultrasound imaging, making it challenging to identify small tumors or subtle tissue abnormalities accurately.

Deconvolution models describe techniques to improve image resolution in ultrasound imaging, by modeling the point-spread-function (PSF) of the imaging system and convolving the image with its inverse \cite{jensen1992deconvolution}.
Deconvolution has been proven useful in various medical tasks such as tissue characterization \cite{alessandriniRestorationFrameworkUltrasonic2011}, prostate cancer detection \cite{maggioPredictiveDeconvolutionHybrid2010}, and microbubble localization \cite{foroozanMicrobubbleLocalizationThreeDimensional2018}.
The basis of this approach is a model of the imaging formation process as a spatially invariant convolution of the PSF with the echogenicity map, which represents the tissue reflectivity function (TRF).
The PSF of an imaging system encapsulates all parameters of the imaging system, such as transducer geometry and frequency response.
By modeling the inverse process of this idealized imaging system via deconvolution, the underlying TRF can be recovered.
In theory, deconvolution eliminates an acquisition-specific influence on the final B-mode image and addresses the resolution challenges associated with a limited bandwidth of the transducer.

In general, deconvolution can be divided into blind (unknown PSF) \cite{goudarziUnifyingApproachInverse2023} and non-blind (known PSF) approaches \cite{dalitzPointSpreadFunctions2015}.
Even though knowing the PSF reduces the solution space, deconvolution is an ill-posed problem where multiple possible solutions exist for one target.
One approach to this problem is to formulate the deconvolution as a Maximum A Posteriori (MAP) estimation, as proposed in~\cite{alessandriniRestorationFrameworkUltrasonic2011}. It provides a structured approach but limits the solution to the given priors and can be noise-sensitive.
Newer methods employ neural networks to execute the deconvolution process \cite{khanUnsupervisedDeconvolutionNeural2020}, demonstrating the potential to recover high-quality images from sparse signals.
Recently, implicit neural representations (INR) have gained attention as a continuous and adaptable mapping function between spatial coordinates and a sensor space \cite{xieNeuralFieldsVisual2022}. At the core, the weights of an MLP are optimized to minimize the difference between regressed and target sensor data. 
INRs have been successfully applied in the medical domain  for sparse-view CT reconstruction \cite{wangMEPNetModelDrivenEquivariant2023, zha2022naf} or for novel view synthesis in ultrasound imaging using a physics-based rendering \cite{wysockiUltraNeRFNeuralRadiance2024}.

Inspired by the recent work, we propose to leverage INRs to learn the continuous underlying echogenicity map from a B-mode image in order to enhance resolution. 
Our approach utilizes a deconvolutional method, based on a modeled PSF, and integrates a fully differentiable rendering pipeline. This allows us to estimate the echogenicity map by comparing predicted and acquired B-mode images. We evaluate the effectiveness of our framework using synthetic, in-vitro, and in-vivo US data.
Our contributions are as follows:
\begin{enumerate}
    \item a novel methodology to retrieve the continuous echogenicity map of soft tissue anatomy by learning its implicit neural representation
    \item a physics-informed, differentiable rendering pipeline that models the ultrasound formation process
\end{enumerate}





\section{Methods}


\subsection{Image Formation Process}

Following \cite{jensenModelPropagationScattering1991, ngModelingUltrasoundImaging2006, zempLinearSystemModels2003}, we model the pulse-echo time series data as a linear system that can be estimated as:
\begin{equation}\label{eqn:temporal_rf}
    y(z, t) = h(z,t) * \gamma(z) + n(t),
\end{equation}
where $y$ is the beamformed radio frequency (RF) data per channel based on time $t$ and distance $z$ from the transducer to the interrogated point. $\gamma$ represents the underlying echogenicity map, $n$ is positive zero-mean Gaussian noise, and $*$ denotes a convolution. 
To remove the dependency on time $t$, we assume a constant speed of sound of $v_{SoS} =1.54$~mm/$\mu$s and can, therefore, write everything in spatial coordinates, given a known signal frequency $f_s$.
This simplifies the equation to:
\begin{equation} \label{eqn:spatial_rf}
    y(z \mid f_{s}, v_{SoS}) = h(z)*\gamma(z) + n(z)
\end{equation}
 To acquire a realistic PSF, we follow \cite{walkerApplicationKspacePulse1998} in their representation of the imaging system as a convolution of the transmit and receive aperture functions multiplied with the axial pulse in $k_{s}$-space. We present the PSF function in the spatial domain as the outer product $\otimes$ of the axial $h_{ax}$ and lateral $h_{lat}$ pressure distribution:

\begin{equation}\label{eq:psf_function}
    h(x', z') = h_{lat}(x') \otimes h_{ax}(z')
\end{equation}

Due to aperture growth and synthetic aperture focusing of modern US machines, we assume an isotropic PSF function in the image, thereby removing the dependence on $z$ in Equation \ref{eqn:spatial_rf}. We furthermore introduce independent axial and lateral spatial variables $x'$ and $z'$.
The lateral pressure distribution at the focal distance $r$ of a beam can be estimated by the convolution of the Fourier transform of the receive and transmit aperture functions \cite{walkerApplicationKspacePulse1998}.
Assuming a uniform apodization function, we model the aperture functions as: 

\begin{equation}
    A_{T}(t, f_s, D) = A_{R}(t, f_s, D) = t \cdot f_x \cdot rect(1/D) 
\end{equation} 
where $A_{T}$ and $A_{R}$ refer to the aperture function of the transmit and the receive signal, respectively, and lateral aperture extent $D$. The evaluation of the lateral signal at an arbitrary focal point $f_x(x) = x/ (\lambda z)$ can be written as:
\begin{equation}
h_{lat}(x'\mid \lambda, D, r) = \mathcal{F}(A_{T}) * \mathcal{F}(A_{R}) =  D^{2} \text{sinc}^2\left( \frac{D x'}{ r \lambda} \right)
\end{equation}
where $\lambda$ is the wavelength and $\mathcal{F}$ represents the Fourier transform. To have a non-negative PSF and remove envelope detection in our pipeline, we model the axial signal as a Gaussian following \cite{dalitzPointSpreadFunctions2015}: 

\begin{equation}
    h_{ax}(z' \mid \sigma_z) = exp(-\frac{z'^2}{2*\sigma_z^2})
\end{equation}

with $\sigma_z$ corresponding to a scaling factor based on the number of cycles for the axial pulse.
Due to the simplification and assumptions, we limit the environment to comply with the first-order Born approximation theorem, similar to \cite{walkerApplicationKspacePulse1998, michailovich2007blind, ngModelingUltrasoundImaging2006, goudarziUnifyingApproachInverse2023}. This implies that an echo of a target only affects the target and does not influence any other backscattering signal \cite{jensenModelPropagationScattering1991}.

\subsection{Proposed Framework}

\begin{figure}[tb!]
\includegraphics[width=\textwidth]{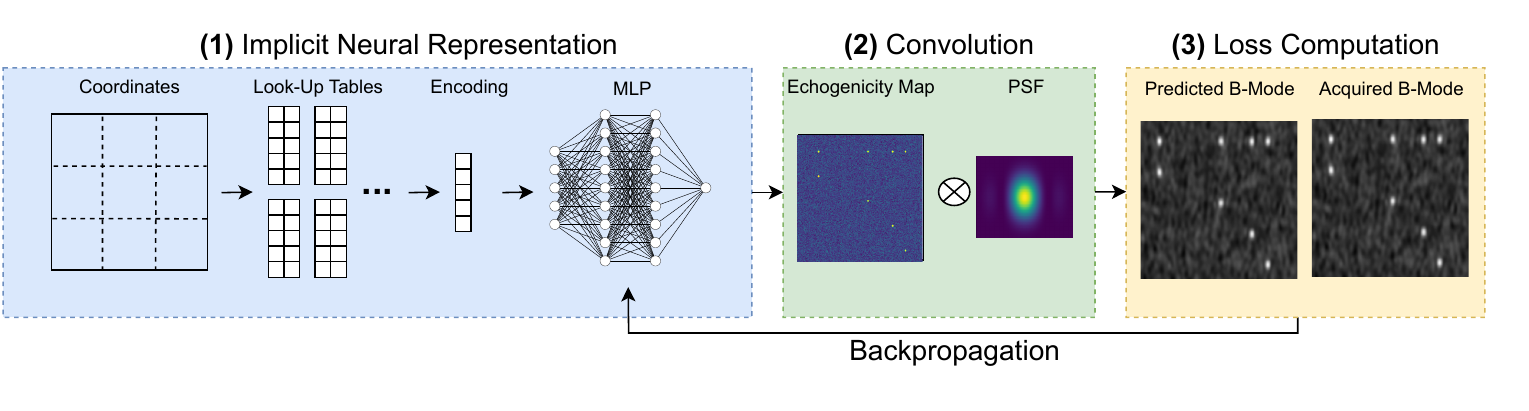}
\caption{General outline of the proposed framework:\textbf{(1)} encoding of the input location based on a multi-resolution look-up table, which forms the input to the MLP (inspired by \cite{mullerInstantNeuralGraphics2022}). The neural network predicts the echogenicity map. \textbf{(2)} convolution of the echogenicity map with the PSF. \textbf{(3)} computation of the loss and backpropagation to update the MLP.} \label{fig1:general_pipeline}
\end{figure}

Our framework consists of three components for approximating the continuous echogenicity map of the tissue, as described in Figure \ref{fig1:general_pipeline}. First, an INR that predicts a scalar echogenicity value $s$ for a given spatial location $x \in \mathbf{R}^{2}$.
Second, a differentiable rendering pipeline convolving a PSF with the predicted echogenicity map. Third, the backpropagation of the computed loss function for updating the parameters in the INR.  

In the first step, the input space is divided into multiple grids of $L$ different resolution levels, where the vertex indices are used to perform a hash-based table lookup. In this table, an $F$-dimensional trainable feature vector is stored. The feature tables have a fixed size of T entries. A bilinear interpolation weighs the inputs based on the respective location in the grid. Finally, the concatenation of the different resolution levels forms the input $\theta \in \mathbf{R}^{LF}$ to the MLP. Based on this encoding, the MLP predicts the echogenicity value $s$. Performing the convolution of the echogenicity map with the PSF, we get the envelope data $e$.
\begin{equation}
    e(x, z) = h(x', z') * s(x, z) 
\end{equation}

Mimicking the image formation process of B-mode images, we perform log compression on the envelope data to get the B-mode image in decibels. To constrain the INR, we include three different loss functions that can be separated into regularizing the output and regularizing the echogenicity map.\begin{equation}
    \mathcal{L} = \lambda \text{SSIM}(I', I) + (1-\lambda) \sum_{i \in I} \left( I'(i) - I(i) \right)^2 + \epsilon TV(S)
\end{equation} 
Similar to \cite{wysockiUltraNeRFNeuralRadiance2024}, we employ SSIM and L2 loss as a similarity measurement between the predicted Image I' and its acquired counterpart I. Additionally, we use a total variation loss for regularizing the predicted echogenicity map $S$. This has been shown to be useful in the regularization of deconvolutional algorithms \cite{dalitzPointSpreadFunctions2015} and enhanced results for us as well. We empirically determined that the best results were acquired by setting $\lambda = 0.5$ and $\epsilon = 1e-4$.



\subsection{Richardson-Lucy Algorithm}

As a comparison, we use the Richardson-Lucy algorithm \cite{lucyIterativeTechniqueRectification1974}, which has been shown to be efficient for ultrasound deconvolution \cite{dalitzPointSpreadFunctions2015}. The Richardson-Lucy algorithm is an iterative algorithm that refines image $f$ with each step $n$ by applying: 
\begin{equation}
    f_{n+1} = \left( \frac{d}{h * f_n} * \bar{h} \right) \cdot f_n
\end{equation}
where $d$ represents the target image. $h$ and $\bar{h}$ denote the PSF and the complex conjugate of the PSF, respectively.

\subsection{Data}
\subsubsection{In Silico Data:}
To compare our method with the baseline, we use the general-purpose ultrasound phantom from CIRS (model 054GS).
For synthetic and real data evaluation, we reconstructed the structure of the CIRS phantom based on available measurements\footnote{https://www.cirsinc.com/wp-content/uploads/2020/12/054GS-UG-062119.pdf} and estimated unknown parameters to visually match the outline of the acquired b-mode of the phantom.

\textit{Axial and Lateral Resolution Targets:}
We use the close proximity targets on the CIRS phantom to evaluate the axial and lateral resolution capability of our approach. In total, 12 nylon wires are spread out at a depth of 3 cm. The wires have a diameter of 80 microns, and the closest wires are 0.25 mm apart from each other.

\textit{Cylindrical Inclusions:} 
The grayscale targets, located at a depth similar to that of the resolution targets, consist of six cylindrical inclusions aligned side by side, each measuring 8 mm in diameter. We visualize three central cylinders with a contrast of (+6, +3, -3) in decibels in comparison to the background.

\textit{Synthetic Image Formation Process: }In order to create the echogenicity map, we take the structures and the mean intensities based on the decibels defined above and then apply Rayleigh distribution-based variations to the different labels. The scale parameter $\sigma$ is calculated based on the relation between the mean $\mu$ of the distribution $f(x \mid \sigma)$ and the scale given by $\mu(f(x\mid \sigma)) = \sigma \sqrt{\frac{\pi}{2}}$.
\begin{equation}
    f(x \mid \sigma) = \frac{x}{\sigma^2} e^{-x^2/(2\sigma^2)}, \quad x \geq 0,
\end{equation}

After the convolution of the generated echogenicity map with a PSF, we perform log compression on the data to get the synthetic B-mode image.

\subsubsection{In-Vitro and In-Vivo:}

Data were acquired using a Siemens Juniper Acuson ultrasound system with a 12L3 linear probe. Images were retrieved from the ultrasound machine using a frame grabber. To eliminate distortions, the phantom was submerged in a water bath, and all post-processing features like tissue harmonic imaging were disabled to produce unprocessed B-mode images. For the in-vivo study, a B-mode image of a healthy volunteer's carotid artery was obtained under an approved institutional review protocol.

\section{Experiments \& Results}

\subsection{Implementation Details}
The MLP and the learnable feature tables were written in Pytorch. We achieved the best ratio of performance and accuracy when setting the number of the learnable parameter table size to $2^{22}$ using only a single-dimensional feature per entry and $15$ resolution levels. Using the learnable tables, the MLP can be rather small, so we set it to 2 layers with 64 neurons as proposed by \cite{mullerInstantNeuralGraphics2022}. Similarly, for optimization, we used Adam as an optimizer with a learning rate of 0.01. Due to the possibility that the pixel resolution may not be sufficiently high, we incorporated oversampling of the spatial grid to meet the Nyquist criteria. To improve continuity, we implemented random sampling between grid points.  Convergence was usually reached after 5000 iterations. All the training was done on a single workstation with an NVIDIA RTX 4070 Ti.  The code for this work can be found on GitHub\footnote{https://github.com/Felixduelmer/phocus}.

\subsection{Synthetic Dataset}
\begin{figure}[h] 
    \includegraphics[width=\textwidth]{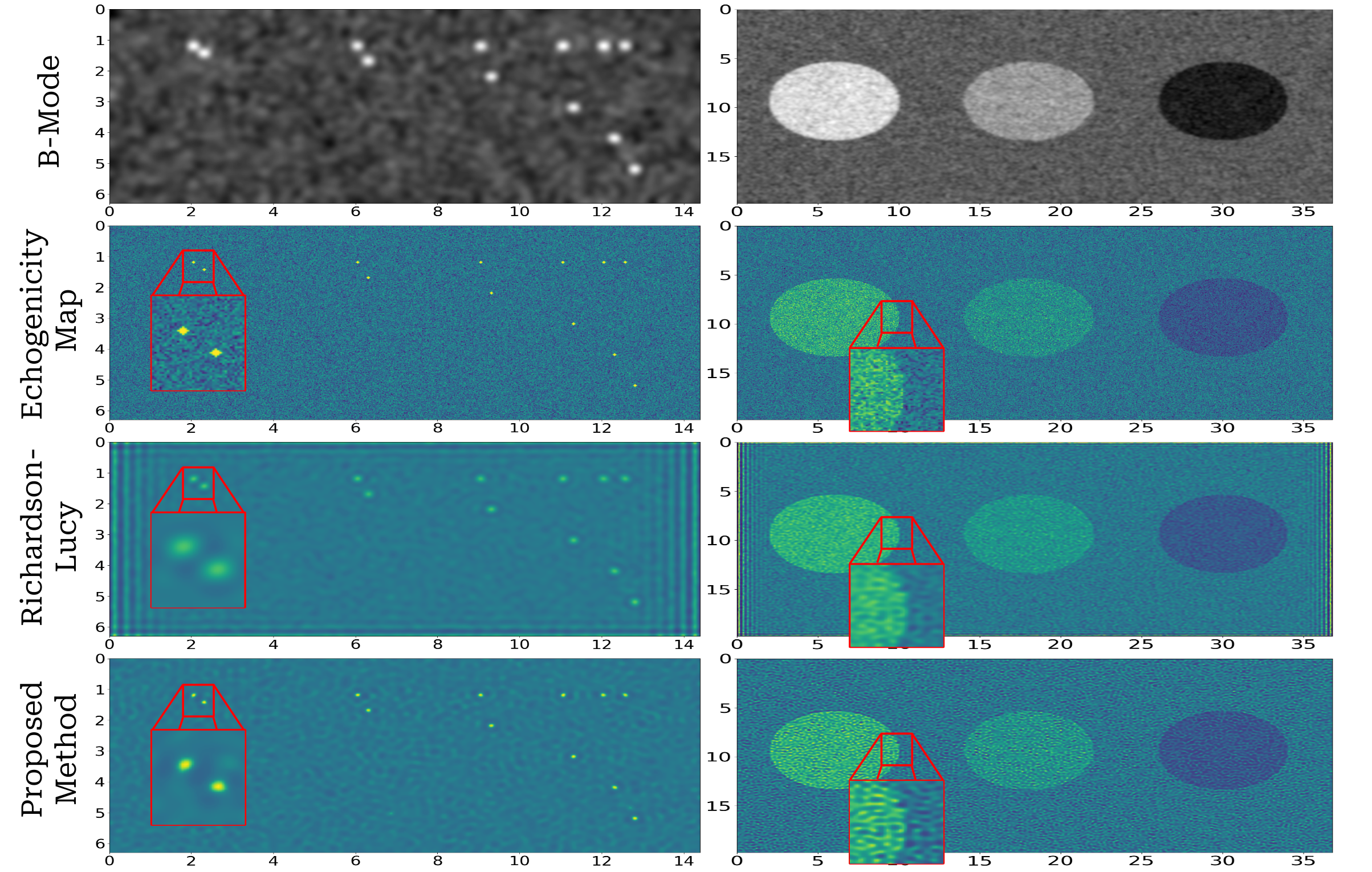}
    \caption{The left column represents the axial and lateral resolution targets, and the right column displays the cylindrical inclusions.} \label{fig:synthetic_comparison}
\end{figure} 

In Figure \ref{fig:synthetic_comparison}, we compare the ground truth echogenicity map and the predicted ones from the Richardson-Lucy algorithm and our proposed methodology. As the Richardson-Lucy algorithm predicts the echogenicity map directly in log-compressed space, we perform log compression on our prediction and on the ground truth map. Figure \ref{fig:synthetic_comparison} highlights, within the enlarged red bounding boxes, that the proposed method generates high-resolution images. Unlike the Richardson-Lucy algorithm, our approach delivers sharper edges and achieves clearer target separation.

\begin{table}[htb!]
\centering
\caption{Quantitative evaluation of the deconvolution algorithms}
\label{tab:correlation}
\begin{tabular*}{\textwidth}{@{\extracolsep{\fill}}lcccc}
\toprule
                 & \multicolumn{2}{c}{Cylindrical Inclusions} & \multicolumn{2}{c}{Wire Targets} \\ 
                 \cmidrule(r){2-5}
                 & PSNR          & SSIM     &    PSNR    & SSIM          \\ 
                 \midrule
Richardson-Lucy  & 16.89         & 0.21     &    17.35   & 0.06          \\
Proposed Method  & \textbf{17.85}& \textbf{0.29} &    \textbf{17.85}   & \textbf{0.07} \\ \bottomrule
\end{tabular*}
\end{table}

Accordingly, we observe better values for Peak Signal-to-Noise Ratio (PSNR) and Structural Similarity Index (SSIM), when comparing the ground truth maps with the predicted echogenicity map (see Table \ref{tab:correlation}).

\subsection{Wire Target In-Vitro}
\begin{figure}[htb!]
    \centering
    \includegraphics[width=\textwidth]{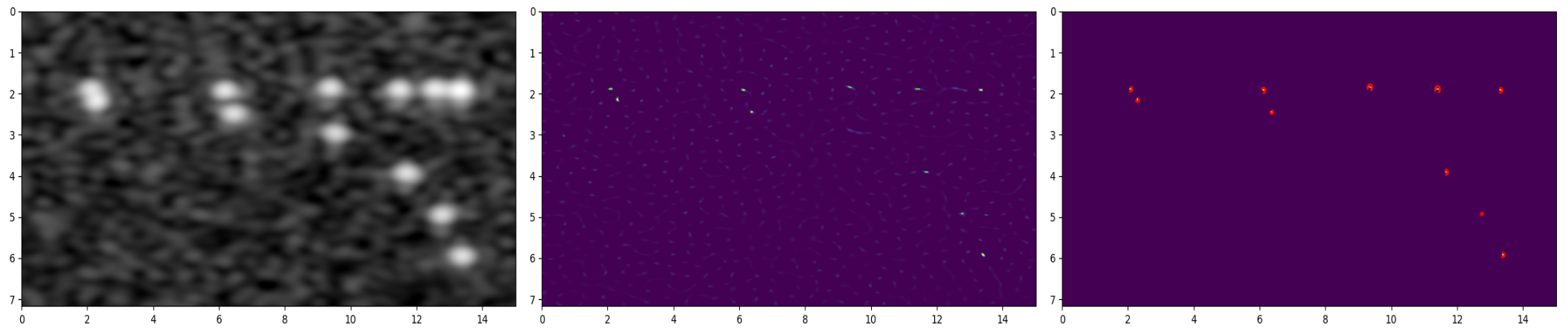}
    \caption{Predicted echogenicity map (center) of the B-mode image (left). The right side shows the result of the minimum enclosing circles' algorithm on the filtered and clustered image.}
    \label{fig:real_deconvolution}
\end{figure}
Following the synthetic results we evaluate the feasibility of our approach on real B-mode data. In order to obtain the PSF, we performed a grid search by varying the f-number (ratio of the focal distance to the diameter of the aperture) from 1.0 to 4.0 in steps of 0.5 and the number of cycles for the transmit pulse from 1 to 5 and selected the best results. As can be seen in Figure \ref{fig:CIRS_wires}, the targets are notably smaller and are clearly separable. For quantitative evaluation, we perform an intensity thresholding at 20 percent, filter for noise, and then perform a clustering. Consequently, we calculate the cluster centers and the respective radius for each cluster. Following this process, we end up with 10 of the 12 expected targets, a mean radius of 0.053 mm, and a standard deviation of 0.01 mm which is close to the expected 0.04 mm value.

\subsection{Carotid Data In-Vivo}

\begin{figure}[htb]
    \centering
    \includegraphics[width=\textwidth]{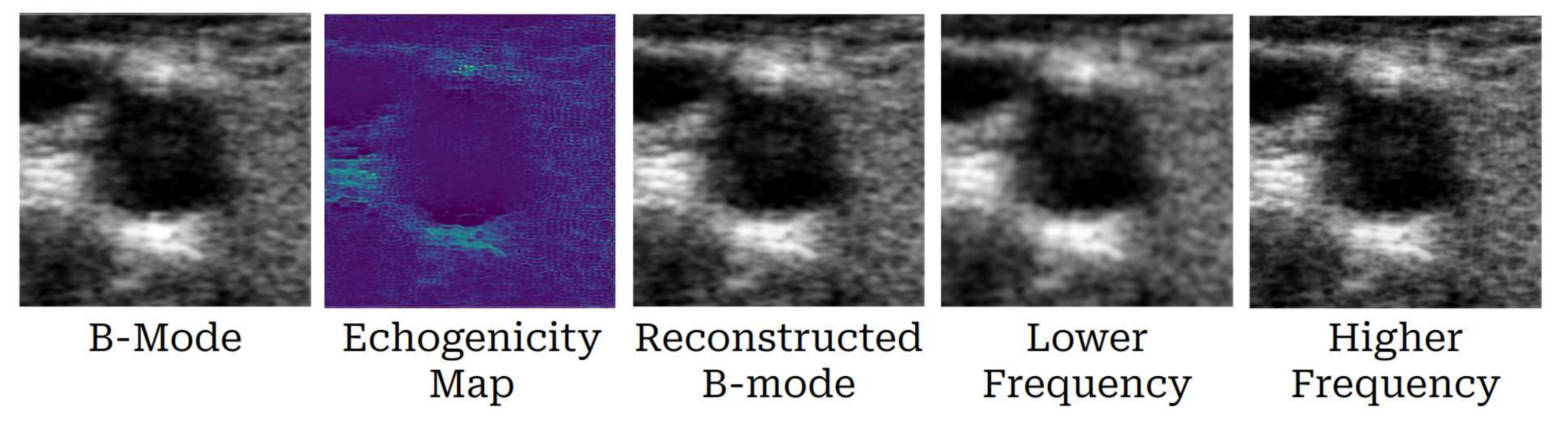}
    \caption{In-Vivo carotid data: the acquired B-mode image with 8 MHz central frequency, the predicted echogenicity map, and the reconstructed B-Mode image at the same frequency, at 6 MHz, and at 10 MHz }
    \label{fig:CIRS_wires}
\end{figure}

Lastly, we demonstrate our approach on the carotid of a volunteer. As can be seen in Figure \ref{fig:CIRS_wires}, we can reconstruct the B-mode image with the predicted echogenicity map. Additionally, the intima layer in the carotid is more clearly visible on the echogenicity map. Evaluating the predicted map, we reconstruct the B-mode image with a baseband and harmonic frequency to display the generalization ability of our proposed method. In this way, we show a tighter speckle and sharper resolution. 

\section{Discussion \& Conclusion}


We present a framework for enhancing ultrasound imaging resolution, focusing on the integration of physical sound propagation principles with advanced computational techniques on commonly available B-mode images. Unlike conventional approaches that work on RF data, our work introduces a methodology for retrieving the echogenicity map directly from the B-mode image. 

However, our approach faces challenges inherent to the ill-posed nature of the problem, such as image alterations from post-processing parameters introduced by the ultrasound machine (e.g., log-compression factor). Additionally, determining the correct PSF through grid search is computationally expensive. A learnable PSF could, therefore, simplify the application of our approach in the future.

In conclusion, this work introduces a novel methodology to retrieve the continuous echogenicity map by learning its implicit neural representation based on a differentiable rendering pipeline that models the ultrasound formation process. The presented approach enhances the quality of ultrasound imaging and can open the door for future improvements in medical diagnostics.
\bibliographystyle{splncs04}
\bibliography{us_learner.bib}

\begin{thebibliography}{10}
\providecommand{\url}[1]{\texttt{#1}}
\providecommand{\urlprefix}{URL }
\providecommand{\doi}[1]{https://doi.org/#1}

\bibitem{alessandriniRestorationFrameworkUltrasonic2011}
Alessandrini, M., Maggio, S., Por{\'e}e, J., De~Marchi, L., Speciale, N., Franceschini, E., Bernard, O., Basset, O.: A restoration framework for ultrasonic tissue characterization. IEEE transactions on ultrasonics, ferroelectrics, and frequency control  \textbf{58}(11),  2344--2360 (2011)

\bibitem{dalitzPointSpreadFunctions2015}
Dalitz, C., Pohle-Frohlich, R., Michalk, T.: Point spread functions and deconvolution of ultrasonic images. IEEE transactions on ultrasonics, ferroelectrics, and frequency control  \textbf{62}(3),  531--544 (2015)

\bibitem{foroozanMicrobubbleLocalizationThreeDimensional2018}
Foroozan, F., O’Reilly, M.A., Hynynen, K.: Microbubble localization for three-dimensional superresolution ultrasound imaging using curve fitting and deconvolution methods. IEEE Transactions on Biomedical Engineering  \textbf{65}(12),  2692--2703 (2018)

\bibitem{goudarziUnifyingApproachInverse2023}
Goudarzi, S., Basarab, A., Rivaz, H.: A unifying approach to inverse problems of ultrasound beamforming and deconvolution. IEEE Transactions on Computational Imaging  \textbf{9},  197--209 (2023)

\bibitem{jensenModelPropagationScattering1991}
Jensen, J.A.: A model for the propagation and scattering of ultrasound in tissue. The Journal of the Acoustical Society of America  \textbf{89}(1),  182--190 (1991)

\bibitem{jensen1992deconvolution}
Jensen, J.A.: Deconvolution of ultrasound images. Ultrasonic imaging  \textbf{14}(1),  1--15 (1992)

\bibitem{khanUnsupervisedDeconvolutionNeural2020}
Khan, S., Huh, J., Ye, J.C.: Unsupervised deconvolution neural network for high quality ultrasound imaging. In: 2020 IEEE International Ultrasonics Symposium (IUS). pp.~1--4. IEEE (2020)

\bibitem{lucyIterativeTechniqueRectification1974}
Lucy, L.B.: An iterative technique for the rectification of observed distributions. Astronomical Journal, Vol. 79, p. 745 (1974)  \textbf{79}, ~745 (1974)

\bibitem{maggioPredictiveDeconvolutionHybrid2010}
Maggio, S., Palladini, A., De~Marchi, L., Alessandrini, M., Speciale, N., Masetti, G.: Predictive deconvolution and hybrid feature selection for computer-aided detection of prostate cancer. IEEE transactions on medical imaging  \textbf{29}(2),  455--464 (2009)

\bibitem{michailovich2007blind}
Michailovich, O., Tannenbaum, A.: Blind deconvolution of medical ultrasound images: A parametric inverse filtering approach. IEEE Transactions on Image Processing  \textbf{16}(12),  3005--3019 (2007)

\bibitem{mullerInstantNeuralGraphics2022}
M{\"u}ller, T., Evans, A., Schied, C., Keller, A.: Instant neural graphics primitives with a multiresolution hash encoding. ACM transactions on graphics (TOG)  \textbf{41}(4),  1--15 (2022)

\bibitem{ngModelingUltrasoundImaging2006}
Ng, J., Prager, R., Kingsbury, N., Treece, G., Gee, A.: Modeling ultrasound imaging as a linear, shift-variant system. ieee transactions on ultrasonics, ferroelectrics, and frequency control  \textbf{53}(3),  549--563 (2006)

\bibitem{walkerApplicationKspacePulse1998}
Walker, W.F., Trahey, G.E.: The application of k-space in pulse echo ultrasound. IEEE transactions on ultrasonics, ferroelectrics, and frequency control  \textbf{45}(3),  541--558 (1998)

\bibitem{wangMEPNetModelDrivenEquivariant2023}
Wang, H., Zhou, M., Wei, D., Li, Y., Zheng, Y.: Mepnet: a model-driven equivariant proximal network for joint sparse-view reconstruction and metal artifact reduction in ct images. In: International Conference on Medical Image Computing and Computer-Assisted Intervention. pp. 109--120. Springer (2023)

\bibitem{wysockiUltraNeRFNeuralRadiance2024}
Wysocki, M., Azampour, M.F., Eilers, C., Busam, B., Salehi, M., Navab, N.: Ultra-nerf: neural radiance fields for ultrasound imaging. In: Medical Imaging with Deep Learning. pp. 382--401. PMLR (2024)

\bibitem{xieNeuralFieldsVisual2022}
Xie, Y., Takikawa, T., Saito, S., Litany, O., Yan, S., Khan, N., Tombari, F., Tompkin, J., Sitzmann, V., Sridhar, S.: Neural fields in visual computing and beyond. In: Computer Graphics Forum. vol.~41, pp. 641--676. Wiley Online Library (2022)

\bibitem{zempLinearSystemModels2003}
Zemp, R.J., Abbey, C.K., Insana, M.F.: Linear system models for ultrasonic imaging: Application to signal statistics. IEEE transactions on ultrasonics, ferroelectrics, and frequency control  \textbf{50}(6),  642--654 (2003)

\bibitem{zha2022naf}
Zha, R., Zhang, Y., Li, H.: Naf: Neural attenuation fields for sparse-view cbct reconstruction. In: International Conference on Medical Image Computing and Computer-Assisted Intervention. pp. 442--452. Springer (2022)

\end{thebibliography}
%




\end{document}